# Understanding the Social Cascading of Geekspeak and the Upshots for Social Cognitive Systems

Michał B. PARADOWSKI[1], Łukasz JONAK[2]

**Abstract.** Barring swarm robotics, a substantial share of current machine-human and machine-machine learning and interaction mechanisms are being developed and fed by results of agent-based computer simulations, game-theoretic models, or robotic experiments based on a dyadic communication pattern. Yet, in real life, humans no less frequently communicate in groups, and gain knowledge and take decisions basing on information cumulatively gleaned from more than one single source. These properties should be taken into consideration in the design of autonomous artificial cognitive systems construed to interact with//learn from more than one contact or 'neighbour'. To this end, significant practical import can be gleaned from research applying strict science methodology to human and social phenomena, e.g. to discovery of realistic creativity potential spans, or the 'exposure thresholds' after which new information could be accepted by a cognitive agent.

The results will be presented of a project analysing the social propagation of neologisms in a microblogging service. From local, low-level interactions and information flows between agents inventing and imitating discrete lexemes we aim to describe the processes of the emergence of more global systemic order and dynamics, using the latest methods of complexity science. Of particular interest is the ability to track those novel linguistic expressions which are *idiosyncratic* to the system (i.e., not used in offline discourse). This allows us to plot the dynamics of the spread of the items *in a closed, hermetic circuit* relative to its structure and size.

We will consider the following issues: 1) how linguistic innovation becomes a norm, 2) that the distribution of the general lexical innovativeness of Internet users scales not like a power law, but a unimodal, 3) that the exposure thresholds characterising users' readiness to adopt new lexemes from their neighbours concentrate at low values, suggesting that—at least in low-stakes scenarios—people are more susceptible to social influence than may erstwhile have been expected, and 4) that, contrary to common expectations, the most popular tags are characterised by high adoption thresholds. Hypotheses will be investigated which may account for the observed phenomena.

Whether in order to mimic them, or to 'enhance' them, parameters gleaned from complexity science approaches to humans' social and humanistic behaviour should subsequently be incorporated as points of reference in the field of robotics and human-machine interaction.

*Why is it that when robots are stored in an empty space, they will group together, rather than stand alone?*
—Dr. Alfred Lanning in *I, Robot* (2004)

## 1. THE ORIGINAL TURING TEST

The year 2012 marks the centenary of the birth of Alan Turing. Yet it is an unjust legacy that to this day few, even in the AI field, seem aware of what Turing originally had in mind when, in his 1950 paper *Computing machinery and intelligence* [29], he introduced his concept of a test examining a machine's ability to exhibit intelligent (or, to be more precise, humanlike) behaviour. For it was not merely a matter of whether a computer could interact with a human in such a way that the interrogator would be deceived into thinking s/he were 'conversing' with another human being. The design of the imitation game was much more subtle: to see whether a computer pretending to be a woman could be more convincing than a man also pretending to be a woman.[3]

## 2. LANGUAGE SIMULATIONS

Regrettably, this necessary proviso of 'other things being equal', so forcefully emphasised in Turing's original scenario, all too often seems to be overlooked in much of current research literature which either has the ambition to serve as input for developing AI, or which could potentially be applied by the field.

For instance, over the past decade much space has been devoted to language simulations, from workshops devoted exclusively to that topic[4] to articles posted on arXiv and published across scientific journals. Many of the papers, devoted to phenomena such as language evolution, language competition, language spread, and semiotic dynamics, were based on regular-lattice *in silico* experiments and as such are glaringly inadequate, especially to the scenery of the 21st century:

- the models take into account only Euclidean relationships (whereas the current telecommunication technology and the global accessibility of mass media mean that more and more of our linguistic input reaches us from afar, and—especially with services such as VoIP calls and social networking sites—spatial proximity can no longer be equated with social proximity);
- are 'static' (while mobility has been a distinctive feature of humankind—but also the animal world—as evidenced by warriors, refugees, missionaries, civil servants, and tradespeople long before the time of the Hanseatic League);
- assume a limited, identical number of 'neighbours' for every agent (4 $\veebar$ 8;[5] first of all, an underestimate, secondly, again unrealistic given that persons vary in terms of the number of their close friends, acquaintances, or relatives – suffice it to think of the growing number of nuclear and patchwork families, the multi-generation families of the not-so-distant past, or the divide in China between urban couples who have had to abide by the one-child policy and the rural countryside where the restriction was not stringently enforced, but where in turn male offspring have often been valued more than female);
- presuppose identical perception of the prestige of a given individual by each of its neighbours (while, again, take a single person known to a group, be it a celebrity or an insider, and their perceived prestige and respect is again going to fluctuate from individual to individual), as well as
- invariant intensity of interactions between the different agents,
- absence of multilingual agents (with a few notable exceptions, e.g. [4]);
- and sometimes more technical issues such as lack of memory effect or zero noise (while noise may be the indicator and initiator of pattern change).

---

[1] Inst. of Applied Linguistics, Univ. of Warsaw, ul. Browarna 8/10, 00-311 Warsaw, Poland. E-mail: `michal.paradowski@uw.edu.pl`.
[2] National Library of Poland, Al. Niepodległości 213, 02-086 Warsaw, Poland. E-mail: `lukasz@jonak.info`.

[3] For an interesting distant but convincing analogy in popular culture, think for instance of the title protagonist of Mankiewicz's 1950 film *All About Eve*, whose faked femininity is more compelling than that of the remaining, genuinely heterosexual heroines.
[4] E.g. the GIACS Workshop on Language Simulations, which took place at the University of Warsaw in the year 2006.

## 3. ALTERNATIVES

This is why there still remains much work in front of the AI circles to move from coarser-grained game-theoretic (e.g. [19]) and agent-based models (e.g. [18]) which not infrequently only manage to capture the initial and final states and the general trend of the phenomena they are purported to describe, towards increasingly accurate and sophisticated work based on the results of rigorous data-driven research and empirical studies that recreate the necessary conditions and parameters as faithfully as possible. One solution is experimental designs involving actual cognitive agents. A new quandary that arises with many designs involving interactions between and learning by (embodied) artificial intelligent agents (for a good overview *cf.* e.g. [26], [27]) is the fact that they are often restricted to dyadic scenarios. This can naturally be justified when the process in which the robots engage is akin to the initial stages of language acquisition in humans, where a baby can conceivably find him-/herself in situations where s/he only interacts with a single caretaker. However, sooner or later the child's interaction becomes more social, with an increased number of input sources and persons against whom linguistic hypotheses can be tested. This calls for research paradigms involving more agents engaging in interactions with the subject under investigation, and luckily more and more robotics teams are moving in this direction.

---
[5] We use the $\veebar$ operator to symbolise exclusive disjunction.

Another, often more time-, cost- and resource-effective alternative is rigorous re-search fuelled by data from genuine human interactions. In the case of linguistic phenomena such as language learning and the uptake of new linguistic expressions, such data can be gleaned by either interviewing each member of a community and additionally verifying their responses against a more objective benchmark such as e.g. standardized test scores (in the case of foreign language acquisition; *cf.* e.g. [21])—admittedly still a time-consuming process, and one laden with the limitations posed by self-assessment—or, an easier way, utilize readily available repositories of user-generated content such as Web 2.0 sites.

The recent information explosion with exponentially increasing vast quantities of rich sources of data, and their widespread availability, has enabled access to huge amounts of data allowing us to investigate human behaviour from new angles. The increased use of the World Wide Web, and the recent availability of user-generated text in particular, provides evident and unprecedented new research opportunities. The data stored on the Internet is virtually unregulated, essentially uncensored, spontaneous, immediately registered, interconnected, and amenable to relatively easy search and exploration with the use of statistical and concordancing tools. Web 2.0 services, with content (co)generated by the users, especially the ones which allow enriching their analyses with information concerning the structure of the connections and interactions between the participating users, are particularly useful for multi-angle explorations of language and social phenomena, such as humans' communicative behaviour. By tapping into the repositories of language data nearly perfectly suited to fine-grained large-scale dynamic linguistic analyses and applying novel, transdisciplinary research methodology, most of the formerly-mentioned limitations can be addressed and bypassed.

## 4. LANGUAGE ON THE INTERNET

Erstwhile research on language evolution and change focused on large time-scales, typically spanning at least several decades. Nowadays, observable changes are taking place much faster. According to [12], a new English word is born roughly every 98 minutes (admittedly a rather overrated estimate owing to methodological problems).

The uptake of novel linguistic creations in the Internet has been commonly believed to reflect the focus of attention in contemporary public discourse (suffice it to recollect the dynamics and main themes of status updates on Twitter following the presidential elections in Iran, Michael Jackson's death, Vancouver Olympic Games, and the recent Oscar gala, last July's L.A. earthquake, the Jasmine Revolution—by some also called the "Internet Revolution"—in Tunisia, the developments in Libya, the 2011 Tōhoku earthquake and tsunami, or ibn Laden's death, see e.g. [11]). However, even where the topics coincide, the proportions in the respective channels of information are divergently different (correlation at a level of a mere .3; e.g. [23], just as television ratings cannot be used to predict online mentions; [20]), just as not infrequently the top stories in the mainstream press are markedly different than those leading on social media platforms (e.g. [24]). The emotive content of comments on different social platforms is also distinctly different ([2], [7]).

## 5. A CASE IN POINT: TAGS AND SOCIAL COORDINATION

In a recent empirical research project [22], we investigated the creation and adoption of tags (metalabels) on the Polish microblogging site Blip.pl (roughly analogous to Twitter), with special emphasis on neological expressions. At the time of the data dump the site had 20k users (with over half logging on daily), with 5.5k users in the giant component[6] (density: 0.003), 110k relations, 38k tags and 720k tagged statuses. The data were analysed in Python.

The intended purpose of tagging systems introduced to various Web 2.0 services was to provide ways of building *ad hoc*, bottom-up, user-generated thematic classifications (or "folksonomies"; [31]) of the content produced or published within those systems. However, the tagging system of Blip became much more than that, as users redefined the meaning and modes of using tags. In the site, tagging is not merely a mechanism for retrospective content classification, but also provides institutional scaffold for on-going communication within the system. From the point of view of *individuals*, using a tag within a status update still provides information about what the update is about, but also implies joining the conversation defined by the tag, and, consequently, subscribing to the rules and conventions governing conversation. In this sense, the system of tags can be thought of as an institution (as sociologically understood), regulating and coordinating social conduct – here, mostly communication. From the *systemic* point of view, tags-institutions define what Blip.pl is about, the meaning of its dynamics, and its culture.

## 6. THE LONG TAIL OF THE BLIP CULTURE

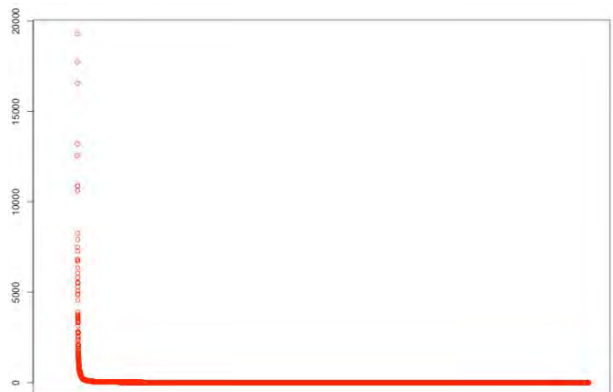

**Figure 1.** Tag popularity distribution in Blip

---
[6] By the term 'giant component' we mean the unique largest connected cluster (subgraph), containing a majority of the entire graph's vertices.

One of the preliminary results obtained from the data analysis carried out concerns tag popularity, whose distribution scales like a power law (Fig. 1), a feature Blip shares with a wide range of natural, technological and socio-cultural phenomena (*cf.* e.g. [5], [17]). Our assumption is that at least a considerable proportion of popular Blip tags constitute the "meaning" and structure of the system, its cultural and institutional establishment, while the long tail consists of more or less contingent representations.

## 7. SOCIAL INFLUENCE AND DIFFUSION

The most important mechanism we are looking for has to do with diffusion of innovation. Diffusion and creation of novelty has been traditionally assumed to be among the most important social processes [8]. In our case, each of Blip's tags, a potential communication coordinator, had been first created by a user, then spread throughout the system with greater or smaller success (see Fig. 2). Some of the most successful, most frequently imitated tags have become Blip's culture and structure.

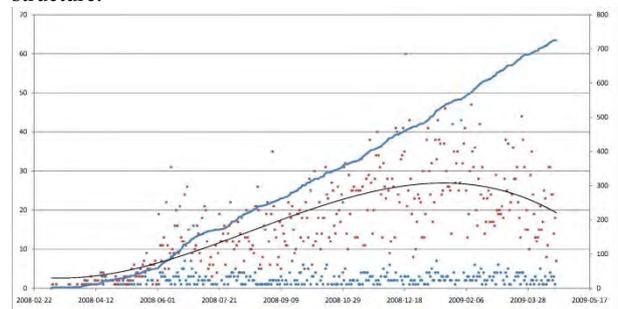

**Figure 2.** Evolution of the popularity of an idiosyncratic tag, relative to system size; abscissæ: time, ordinates left: percentage of saturation; ordinates right: absolute count; blue rhomb dots: first usages; red square dots: subsequent usages; thin black line: subsequent usage trend (multinomial); thick blue line: first usages cumulative

There are a number of theories explaining the mechanisms of diffusion of novelty, and one of our goals is to find out which best accounts for our data. Memetic theory assumes that ideas (here coded as words-tags) are like viruses which "use" the mechanisms of the human mind to reproduce. The most successful reproducers would be those optimally adapted to the environment of the mind – its natural dispositions and the ecosystem of already established ideas ([6], [9]).

The theory of social influence proposes that individual behaviour (including adoption of innovation) is contingent on peer pressure. The threshold model of collective behaviour postulates that a person will adopt a given behaviour only after a certain proportion of the people s/he observes have already done the same. This proportion—the "adoption threshold"—constitutes the individual characteristic of each member of the group [13]. The network version of this theory proposes that an individual ("ego") observes only a fraction of the social system, namely, the alters in his/her ego-network. The exposure of the ego to an innovative idea is hence defined as the proportion of his/her alters/neighbours that had already adopted the relevant innovation by the time concerned, and an individual's adoption threshold is computed as his/her network exposure at the time of adoption [30].

A third point of view is offered by the social learning theory [1], which assumes that innovation or behaviour adoption is a result of a psycho-cognitive process which involves evaluation of other people's behaviour and its consequences. In this case the adoption process is perceived as more reflexive and less automatic than the previous two ([14], [25]).

The preliminary analysis conducted involved calculating thresholds for all tag adoptions (i.e., their *first* usages). We describe the user-tag network with a bipartite graph $G = G(U,X,E)$, where $U$ is the set of users, $X$ is the set of tags, and $E$ represents the edges between users and tags. The user-user network we define using a directed graph $D = D(U,H)$, where $H$ is the set of edges, and $e_{u \to x} \in E$ is an edge connecting user $u$ to tag $x$ added in time $\tau_{u \to x}$. Using this notation, we calculate the (mean) measure of the number of alters (neighbours) who had adopted a given tag before user $u$. We only consider first usages:

$$\beta_u = \frac{\sum_{e_{u \to x} \in E(u)} \frac{A_{(e_{u \to x})}}{H_{(t)}(u)}}{|E(u)|}$$

where:
- $A(e_{u \to x})$ is the number of neighbours of $u$ who are already connected to $x$ at time $\tau_{u \to x}$;
- $H_{(t)}(u)$ is the number of neighbours of $u$ at time $t$;
- $E(u)$ is the total number of (unique) tags used by $u$.

The smoothed distribution of $\beta_u$ is plotted below

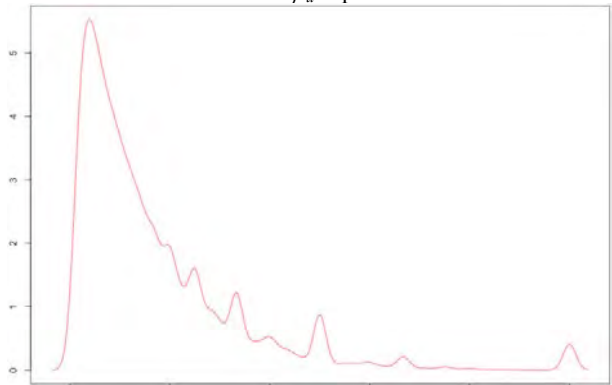

**Figure 3.** Distribution of tag adoption thresholds in Blip

Fig. 3 shows the general characteristics of the innovativeness of Blip's users: the distribution of adoption thresholds based on all instances of tag adoption events. Contrary to Granovetter's assumption of normality of thresholds' distribution in populations, the resultant distribution of adoption thresholds is considerably skewed, with a median of 0.11 and a long tail of higher values[7]. This suggests that the population of Blip's users is generally innovative and/or corroborates the viral model of diffusion over the two alternative mechanisms (social influence and social learning) mentioned earlier. The prevalence of low thresholds indicates that adoption is less contingent on social pressure to adopt, or elaboration of the way the tags are being used by alters, than on individual, cognitive mechanisms of attention and knowledge integration. However, we expect other factors (such as tag and user characteristics) to play an important role as well.

Another finding is the general correlation between tag popularity and adoption threshold. Figure 4 is a scatterplot showing the relationship between the adoption threshold and the general, systemic popularity of a tag at the time of a singular adoption. It show that the more popular tags tend to be adopted at higher values of exposure (which constitutes the adoption threshold of a given user for this tag) than those less popular. This may run contrary to common-sense expectations that the more popular an item is, the more readily it should be adopted. There are two alternative and yet to be verified hypotheses that may account for our observation.

One account supports the social influence approach, and explains the observed relationship by the fact that a lower popularity of a tag *implies* that only people with low adoption thresholds had adopted it by the moment of measurement. The greater popularity of the other tags may simply mean that their diffusion took long enough for people with higher threshold to pick them up. This suggests the classical diffusion process with population division into early adopters and laggards: thresholds rise with tags' popularity because users with lower thresholds had adopted them earlier (when the expressions were not yet popular).

The other hypothesis, corroborating the premises of the social learning theory, postulates that a higher threshold is consistent with later adoption, the adoption lag being needed for observation and evaluation how an innovation is being used by others and works in the social context. The positive correlation between threshold and popularity may stem from the fact that the most popular tags constitute the institutional and cultural structure of the system and so more time is needed for their evaluation, learning and adoption.

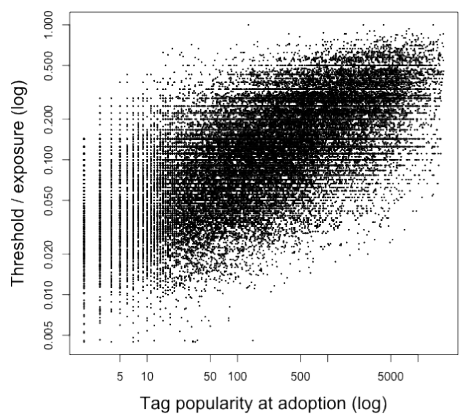

**Figure 4.** Relationship between tag popularity and exposure threshold

There are obviously only a subset of preliminary findings. The next step is to develop a formal model and simulation that will include these factors in explaining diffusion mechanisms in order to gain not only a controlled understanding of their dynamics, but also predictive potential.

---

[7] The "humped" feature of the distribution tail stems from the skewed distribution of the variables used to calculate the threshold values.

## 8. INSIGHTS FOR AI?

Such rigorous data-driven research offers the chance of not only approximating to descriptive adequacy, but also moving beyond explanatory adequacy to approaching principled explanation. Results like the above are conceivably useful to answer questions how a) creative, and b) susceptible to the influence of alters an artificial cognitive agent should optimally be. The established parameters can also be helpful in the development of interactive dialogue systems, HCI, and intelligent machines that acquire knowledge via interaction with other (human and non-human) agents (rather than all their knowledge being put in by their creators). Naturally, the question needs to be posed to what degree observed online behaviour—which may naturally be affected by the medium—can be treated as a realistic proxy for offline behaviour. If we grant the assumption that any difference that may exist is insubstantial, such and similar data-driven research can have practical import for the discipline of artificial (social) intelligence, providing a reference point for at least three aspects of cognitive systems' behaviour:

(i) interaction[8],
(ii) learning, and
(iii) collective intelligence.[9]

Where the agents are expected to pass off as humans, exhibiting performance indistinguishable from that of mankind, e.g. in affective contexts (where, for instance, their task is that of companions), the established data could then be used to emulate human behaviour as closely as possible (bearing in mind the desirability of optimal distinctiveness (*cf.* [3], [15]) and the uncanny valley problem; [16]). In other scenarios, it may be more desirable for the agents to outperform humans[10] (think, for instance, of Deep Blue defeating Kasparov in 1997, or IBM's another wunderkind, Watson, the computer capable of answering natural-language queries, which in February 2011 won the Jeopardy quiz show against two of its all-time human champions). In that case, it is still useful to have a reference point or benchmark. Only subsequently, given the growing sophistication of tools for ABMs, can fine-grained simulations be employed to try to emulate and explain the behaviour observed. This is what Alan Turing would appreciate.[11]

## ACKNOWLEDGMENTS

The authors are supported by a grant from the Polish Association for Social Psychology, Polish Internet Research, and gazeta.pl.

---

[8] And each utilizable cognitive system must be interactive.
[9] Consider e.g. the "wisdom of the crowd" effect ([10], [28]).
[10] To use the words of Gigolo Joe from *A.I.* (2001), "Man made us better at what we do than was ever humanly possible."
[11] Even if computational cognitive systems may be *non*-Turing, with non-terminating computations, interactivity, and non-uniform evolution [32].